\documentclass[10pt,journal,twoside,twocolumn]{article}

\usepackage[a4paper, left=0.5in, right=0.5in, top=0.75in, bottom=0.75in]{geometry}
\usepackage{times}
\usepackage[colorlinks=true,linkcolor=blue]{hyperref}%
\usepackage{amsmath,amssymb,amsfonts}
\usepackage{algorithmic}
\usepackage{graphicx}
\usepackage{xcolor}
\usepackage{subcaption}
\usepackage{multirow}
\usepackage{placeins}
\usepackage{booktabs} 
\usepackage{textcomp}
\usepackage{threeparttable}
\usepackage{makecell} 
\usepackage{cite}
\usepackage{hyperref}
\usepackage{rotating}
\usepackage[font=small,labelfont=bf]{caption}

\usepackage{titling}
\pretitle{\centering\vspace{1em}\rule{\textwidth}{0.5pt}\vspace{1em}\Huge\bfseries}
\posttitle{\par\vspace{1em}\rule{\textwidth}{0.5pt}\vspace{2em}}

\usepackage{sectsty}

\definecolor{darkblue}{rgb}{0,0.541,0.855}%
\sectionfont{\color{darkblue}}
\subsectionfont{\color{darkblue}}
\subsubsectionfont{\color{darkblue}}

\usepackage{fancyhdr}
\date{}

\renewenvironment{abstract}
  {\noindent\textbf{}\par\noindent\ignorespaces}
  {\par\noindent\ignorespacesafterend}

\begin{document}

\title{ \\Unsupervised 4D Flow MRI Velocity Enhancement and Unwrapping Using Divergence-Free Neural Networks}

\author{Javier Bisbal, Julio Sotelo, Hernán Mella, Oliver Welin Odeback, Joaquín Mura, David Marlevi,\\ Junya Matsuda, Kotomi Iwata, Tetsuro Sekine, Cristian Tejos,  and Sergio Uribe
\thanks{This work was supported in part by the National Agency for Research and Development (ANID) / Scholarship Program / DOCTORADO BECAS CHILE/2022 – 21220454; in part by ANID - Millennium Science Initiative Program - ICN2021\_004; in part by Fondecyt 1231535; in part by ANID-FONDECYT Regular 1261167; and in part by Fondecyt de Iniciación en Investigación project No. 11241098.\\
J. Bisbal and C. Tejos are with the Biomedical Imaging Center, Pontificia Universidad Católica de Chile (PUC), Santiago, Chile; with the Department of Electrical Engineering, PUC, Santiago, Chile; and with Millennium Institute for Intelligent Healthcare Engineering, iHEALTH, Santiago, Chile. (e-mails: jebisbal@uc.cl, ctejos@uc.cl).\\
J. Sotelo is with the Departamento de Informática, Universidad Técnica Federico Santa Maria, Santiago, Chile. (e-mail: julio.sotelo@usm.cl).\\
H. Mella is with the School of Electrical Engineering, Pontificia Universidad Católica de Valparaiso (PUCV), Valparaíso, Chile; and with the Center for Interdisciplinary Research in Biomedicine, Biotechnology and Well-Being (CID3B), PUCV, Chile. (e-mail: hernan.mella@pucv.cl)\\
J. Mura is with the Departamento de Ingeniería Mecánica, Universidad Técnica Federico Santa Maria, Santiago, Chile. (e-mail: joaquin.mura@usm.cl).\\
O. Welin Odeback and D. Marlevi are with the Karolinska Institute, Department of Molecular Medicine and Surgery, Stockholm, Sweden. (e-mails: {oliver.welin.odeback@ki.se, david.marlevi}@ki.se)\\
T. Sekine and K. Iwata are with the Department of Radiology, Nippon Medical School, 1-1-5 Sendagi, Bunkyo-ku, Tokyo, Japan. (e-mails: tetsuro.sekine@gmail.com, kotomi-iwata@nms.ac.jp)\\
J. Matsuda is with the Department of Cardiovascular Medicine, Nippon Medical School, 1-1-5 Sendagi, Bunkyo-ku, Tokyo, Japan. (e-mail: jun1984087@nms.ac.jp)\\
S. Uribe and H. Mella are with the Department of Medical Imaging and Radiation Sciences, School of Primary and Allied Health Care. Faculty of Medicine, Nursing and Health Sciences. Monash University, Melbourne, Australia. (e-mail: sergio.uribe@monash.edu).}}

\maketitle

\begin{abstract}
\textbf{This work introduces an unsupervised Divergence and Aliasing-Free neural network (DAF-FlowNet) for  4D Flow Magnetic Resonance Imaging (4D Flow MRI) that jointly enhances noisy velocity fields and corrects phase wrapping artifacts. DAF-FlowNet parameterizes velocities as the curl of a vector potential, enforcing mass conservation by construction and avoiding explicit divergence-penalty tuning. A cosine data-consistency loss enables simultaneous denoising and unwrapping from wrapped phase images. On synthetic aortic 4D Flow MRI generated from computational fluid dynamics, DAF-FlowNet achieved lower errors than existing techniques (up to 11\% lower velocity normalized root mean square error, 11\% lower directional error, and 44\% lower divergence relative to the best-performing alternative across noise levels), with robustness to moderate segmentation perturbations. For unwrapping, at peak velocity/velocity-encoding ratios of 1.4 and 2.1, DAF-FlowNet achieved 0.18\% and 5.2\% residual wrapped voxels, representing reductions of 72\% and 18\% relative to the best alternative method, respectively. In scenarios with both noise and aliasing, the proposed single-stage formulation outperformed a state-of-the-art sequential pipeline (up to 15\% lower velocity normalized root mean square error, 11\% lower directional error, and 28\% lower divergence). Across 10 hypertrophic cardiomyopathy patient datasets, DAF-FlowNet preserved fine-scale flow features, corrected aliased regions, and improved internal flow consistency, as indicated by reduced inter-plane flow bias in aortic and pulmonary mass-conservation analyses recommended by the 4D Flow MRI consensus guidelines. These results support DAF-FlowNet as a framework that unifies velocity enhancement and phase unwrapping to improve the reliability of cardiovascular 4D Flow MRI.}
\end{abstract}

\section{Introduction}
\label{sec:introduction}

Time-resolved 3D phase-contrast magnetic resonance imaging (4D Flow MRI) provides volumetric, time-resolved velocity fields, enabling comprehensive \textit{in vivo} characterization of cardiovascular function by quantifying several hemodynamic parameters \cite{markl_4d_2012, sotelo_fully_2022}. However, its clinical utility is limited by low velocity-to-noise ratio (VNR), poor spatiotemporal resolution, and phase wrapping artifacts \cite{jiang_quantifying_2015}, which can hinder the estimation of advanced hemodynamic parameters such as wall shear stress or pressure changes \cite{ko_novel_2019,petersson_assessment_2012, marlevi_estimation_2019}.

One strategy to improve velocity estimates is to constrain 4D Flow MRI data using physical principles, such as mass conservation, which implies that the estimated velocity fields must be divergence-free. This means that there is no net outflow or inflow through any small closed volume in the fluid. Existing algorithms fall into three groups: (i) weighted least-squares formulations \cite{bostan_improved_2014,mura_enhancing_2016,zhang_divergence-free_2021}, (ii) projection-based methods, which enhance the flow by projecting it onto a divergence-free basis \cite{deriaz_divergence-free_2006,busch_reconstruction_2013, ong_robust_2015}, and (iii) physics-informed deep learning formulations \cite{ferdian_4dflownet_2020,rutkowski2021enhancement,fathi_super-resolution_2020,kalajahi2025input,shone2023deep}.

Weighted least-squares formulations typically minimize a cost function that includes both a data fidelity term and a physics regularization term to obtain a smoothed or ``relaxed" divergence-free solution \cite{bostan_improved_2014,mura_enhancing_2016, zhang_divergence-free_2021}. However, the balance between data-fidelity terms and the divergence constraint may become suboptimal when applied to new datasets. 


Projection-based methods denoise 4D Flow MRI by representing velocity fields with divergence-free basis functions, such as radial basis functions (RBFs) or divergence-free wavelets (DFWs) \cite{busch_reconstruction_2013,ong_robust_2015}. Although projection-based methods can enforce the divergence-free constraint within the vessel lumen, they have their own limitations. For RBFs, the support size must be carefully selected to prevent flow outside the vessel region from propagating errors across the lumen. In the case of DFWs, Ong et al. \cite{ong_robust_2015} used  \textit{SureShrink} soft-thresholding \cite{donoho_adapting_1995} in the non-divergence-free components to capture important velocity components, especially at the boundaries. However, in some cases, \textit{SureShrink} may be too conservative, and a manually tuned threshold may be required.

CNN-based approaches have also been proposed to enhance and super-resolve 4D Flow MRI by learning mappings from low-quality to improved velocity fields using supervised training on synthetic CFD-derived data \cite{ferdian_4dflownet_2020,rutkowski2021enhancement}. These methods can improve denoising and spatial resolution, but their performance depends on the realism of the training data, and generalization across anatomies, acquisition protocols, and flow regimes remains challenging. 




Furthermore, 4D Flow MRI is susceptible to phase wrapping artifacts when velocities exceed the velocity encoding parameter (VENC), e.g., in aortic valve stenosis or aortic coarctation \cite{garcia_role_2019,hom_velocity-encoded_2008}, due to their large peak velocities. In these cases, unwrapping steps are required but often fail to correct all affected voxels, particularly at low VNR. Most methods for velocity enhancement based on divergence-free principles assume that the signal is unaffected by wrapping artifacts. Residual wrapping can therefore introduce large local errors that propagate to subsequent analysis.

Recently, physics-informed neural networks (PINNs) have demonstrated the ability to jointly denoise, unwrap, and super-resolve 4D Flow MRI \cite{fathi_super-resolution_2020}. However, their performance often depends on careful hyperparameter tuning, which limits generalization. Input-parameterized PINNs (IP-PINNs) were introduced to improve the transferability of PINNs by encoding complex synthetic 4D Flow MRI data into latent vectors, thereby reducing training time and improving generalization on unseen domains \cite{kalajahi2025input}. However, IP-PINN is limited by its 4D Flow MRI encoding approximation, which relies only on temporal averaging and k-space truncation.

Although recent PINN-based approaches have shown that denoising, artifact mitigation, and super-resolution can be addressed within neural formulations, their validation has been restricted to synthetic datasets and a single in-vitro aneurysm experiment for PINNs \cite{fathi_super-resolution_2020} and only synthetic datasets for IP-PINNs \cite{kalajahi2025input}. A recent review of PINNs in hemodynamics concluded that clinical translation remains constrained by limited validation, robustness, and scalability, with current successes still concentrated in controlled, small-scale settings \cite{yu2026review}.


One approach to improve robustness and generalization is to incorporate the divergence-free condition directly into the neural network architecture \cite{richter2022neural, garay2024physics}. Garay et al. \cite{garay2024physics} employed a vector potential formulation to enforce zero divergence in PINNs for parameter estimation of reduced-order blood flow models, but this approach has not yet been validated for physically constrained 4D Flow MRI reconstruction. In the current work, we propose a divergence- and aliasing-free neural network architecture (DAF-FlowNet) that estimates a vector potential whose curl yields a divergence-free velocity field, inherently satisfying mass conservation. To handle phase wrapping, we trained our network using a cosine-based loss function \cite{carrillo_optimal_2019}, enabling simultaneous denoising and unwrapping. Unlike prior PINN-based approaches for 4D Flow MRI, DAF-FlowNet enforces divergence-freeness by construction rather than through a soft penalty term, avoiding explicit tuning of divergence weights. 

The main contributions of this work are: (1) a unified unsupervised framework for joint 4D Flow MRI velocity enhancement and phase unwrapping based on a divergence-free vector-potential formulation; (2) improved denoising, divergence minimization, and unwrapping performance at medium and low VENC, relative to competing methods on synthetic data; and (3) an in-vivo evaluation on a cardiovascular patient cohort, demonstrating improved mass conservation in both the aortic and pulmonary circulations without extensive per-patient hyperparameter tuning.

\section{Methods}

\subsection{Neural network formulation}

\subsubsection{Vector potential}
Assuming that blood is an incompressible fluid, we can model the velocity field $\mathbf{V}$ as the curl of a vector potential $\mathbf{\Phi}$:

\begin{align}
\begin{split}
\mathbf{V}&= \nabla \times \mathbf{\Phi} \\ 
&=\left(\frac{\partial \Phi_z}{\partial y}-\frac{\partial \Phi_y}{\partial z}, \frac{\partial \Phi_x}{\partial z}-\frac{\partial \Phi_z}{\partial x}, \frac{\partial \Phi_y}{\partial x}-\frac{\partial \Phi_x}{\partial y}\right) \\
&=(u, v, w)
\end{split}
\label{eq1}
\end{align}

\noindent
inherently satisfying the divergence-free condition: 

\begin{equation}
    \nabla \cdot \mathbf{V}=\nabla \cdot(\nabla \times \mathbf{\Phi})=0
\end{equation}

\subsubsection{Implementation design}

For each timeframe of the 4D Flow MRI dataset, we employ a  fully connected neural network with GELU activation functions \cite{hendrycks2016gaussian}. The network takes the spatial coordinates of the 4D Flow MRI data as input and estimates the vector potential $\mathbf{\Phi}$ as defined in \autoref{eq1}. The velocity field $\mathbf{V}$ is then computed using automatic differentiation throughout the network. Input coordinates are sampled within the fluid region using a mask derived from the 3D Phase Contrast Magnetic Resonance Angiography (PC-MRA) image \cite{markl_comprehensive_2011}:

\begin{equation}
    \text {3D PC-MRA}(\mathbf{r})=\sqrt{\frac{1}{N} \sum_{t=1}^N \left( M_t^2(\mathbf{r}) \sum_{i=x, y, z} \mathbf{V}_{t, i}^2(\mathbf{r})\right)} 
\end{equation}

\noindent
with $M_t$ the magnitude of the 4D Flow MRI signal at timeframe $t$, $V_{t, i}$ the component $i$ of the velocity at timeframe $t$, $\mathbf{r}$ the three-dimensional spatial location, and N the number of timeframes. The fluid region mask was derived using a thresholding tool from a publicly available MATLAB toolbox\footnote{https://github.com/JulioSoteloParraguez/4D-Flow-Matlab-Toolbox} \cite{sotelo20163d}; specifically, we applied an intensity threshold to suppress background signals and retained the largest connected components.

Coordinates are normalized between 0 and 1 based on their minimum and maximum values to standardize the neural network inputs. To mitigate spectral bias and improve the network's ability to learn high-frequency features, we encoded the input coordinates using Fourier features \cite{tancik_fourier_2020}. Specifically, the input coordinates $\mathbf{r}$ were transformed using a random Fourier mapping $\gamma(\mathbf{r}) = [\cos(2\pi\mathbf{B}\mathbf{r}), \sin(2\pi\mathbf{B}\mathbf{r})]$, where each entry of $\mathbf{B} \in \mathbb{R}^{m \times 3}$ is sampled from $\mathcal{N}\left(0, \sigma^2\right)$ with a standard deviation $\sigma$, also known as Fourier scale, controlling the bandwidth of the encoded features, and $m$ the size of the Fourier mapping.

To enhance the velocities by leveraging the divergence-free condition, even in the presence of wrapping artifacts, we trained our network with a loss function comprising a cosine-based data consistency term \cite{carrillo_optimal_2019} and a no-slip boundary condition:

\begin{equation}
\mathcal{L}=\mathcal{L}_{\text{d}}+ \mathcal{L}_{\text{ns}}
\label{eq:loss_total}
\end{equation}
where 
\begin{equation}
\mathcal{L}_{\text{d}} :=
\sum_{\substack{i\in\{x,y,z\}\\ \mathbf{r}\in \Omega_f}}
1-\cos\!\left(\frac{\pi}{\mathrm{VENC}}
\left(\mathbf{V}_i^{\text{pred}}(\mathbf{r})-\mathbf{V}_i^{\text{orig}}(\mathbf{r})\right)\right)
\end{equation}
and
\begin{equation}
    \mathcal{L}_{\text{ns}}:=\sum_{\mathbf{r}\in\Omega_w}\left\|\mathbf{V}^{\text{pred}}(\mathbf{r})\right\|_2^2.
\end{equation}
\noindent
Here, $\Omega_f$ denotes the set of voxel coordinates inside the segmented lumen (fluid domain) and $\Omega_w$ denotes the set of voxel coordinates of the boundary. The data term $\mathcal{L}_d$ is computed over $\Omega_f$ using the measured noisy and potentially wrapped velocities $\mathbf{V}_i^{\text{orig}}$ for each encoding direction $i\in\{x,y,z\}$. The no-slip term $\mathcal{L}_{\mathrm{ns}}$ is computed over $\Omega_w$ to penalize non-zero predicted velocities at the vessel wall. The wall region $\Omega_w$ was defined as the set of voxel coordinates on the outer boundary of the segmented lumen, obtained by subtracting $\Omega_f$ from its morphological dilation using a structuring element with 6-connectivity.








\subsubsection{Architecture design} \label{ArchitectureDesign}

We performed a grid search to identify the optimal network hyperparameters, including depth, width, the size of the Fourier features embedding, and the Fourier scale $\sigma$. In addition, we conducted a separate sweep over $\sigma$, with all other hyperparameters fixed. For all the experiments, we considered the learning rate ($10^{-3}$ with exponential decay by a factor of 0.9 every 500 epochs) and the number of epochs (3500) as fixed hyperparameters. We also report hyperparameter importance using the Weights \& Biases parameter-importance analysis \cite{wandb}, which trains a random-forest surrogate model to estimate the contribution of each hyperparameter to model performance following the functional ANOVA framework of \cite{hutter2014efficient}. All hyperparameters, including values for grid search, selected, and fixed hyperparameters, are shown in \autoref{tab:network_hyperparameters}.

\begin{table}[t]
\centering
\small
\setlength{\tabcolsep}{5pt}
\caption{Network hyperparameters: grid-search ranges, selected optimal values, and fixed training hyperparameters are reported.}
\label{tab:network_hyperparameters}
\begin{tabular}{lll}
\toprule
Hyperparameter & Grid-search values & Selected / fixed \\
\midrule
Depth & $\{3,4,5,6\}$ & $5$ \\
Width & $\{100,200,300,400\}$ & $200$ \\
Embedding size & $\{64,128,256\}$ & $256$ \\
Fourier scale ($\sigma$) & $\{0.05,0.1,1.0,3.0\}$ & $1.0$ \\
\midrule
Learning rate & $10^{-3}$ with exp. decay & fixed \\
Epochs & $3500$ & fixed \\
\bottomrule
\end{tabular}
\end{table}

For in-vivo cases, the Fourier scale $\sigma$ was adjusted using a geometric scaling factor ($\tau$) to account for relative size and resolution differences between in-vivo and simulation domains

\begin{equation}
    \tau = \prod_{i \in \{x,y,z\}} \left(\frac{s_{\text{sim}}^i}{s_{\text{in-vivo}}^i}\right)\left(\frac{R_{\text{in-vivo}}^i}{R_{\text{sim}}^i}\right), \ \
    \sigma_{\text{in-vivo}} = \tau \, \sigma_{\text{sim}}. 
    \label{eq:scaling_factor}
\end{equation}

\noindent
Here, $s^i$ represents the voxel spacing in dimension $i$, and $R^i$ denotes the spatial extent of the masked region in dimension $i$. The size-ratio term $\left(R_{\text{in-vivo}}^i / R_{\text{sim}}^i\right)$ accounts for differences in normalized coordinates caused by changes in domain extent, so that the range of sampled frequencies of $\mathbf{B}$ adapts to different size ratios. The resolution term $\left(s_{\text{sim}}^i / s_{\text{in-vivo}}^i\right)$ adjusts the frequency range according to voxel spacing, increasing it for finer resolutions and decreasing it for coarser ones.





\begin{figure*}[!ht]
    \centering
    \includegraphics[width=\textwidth,]{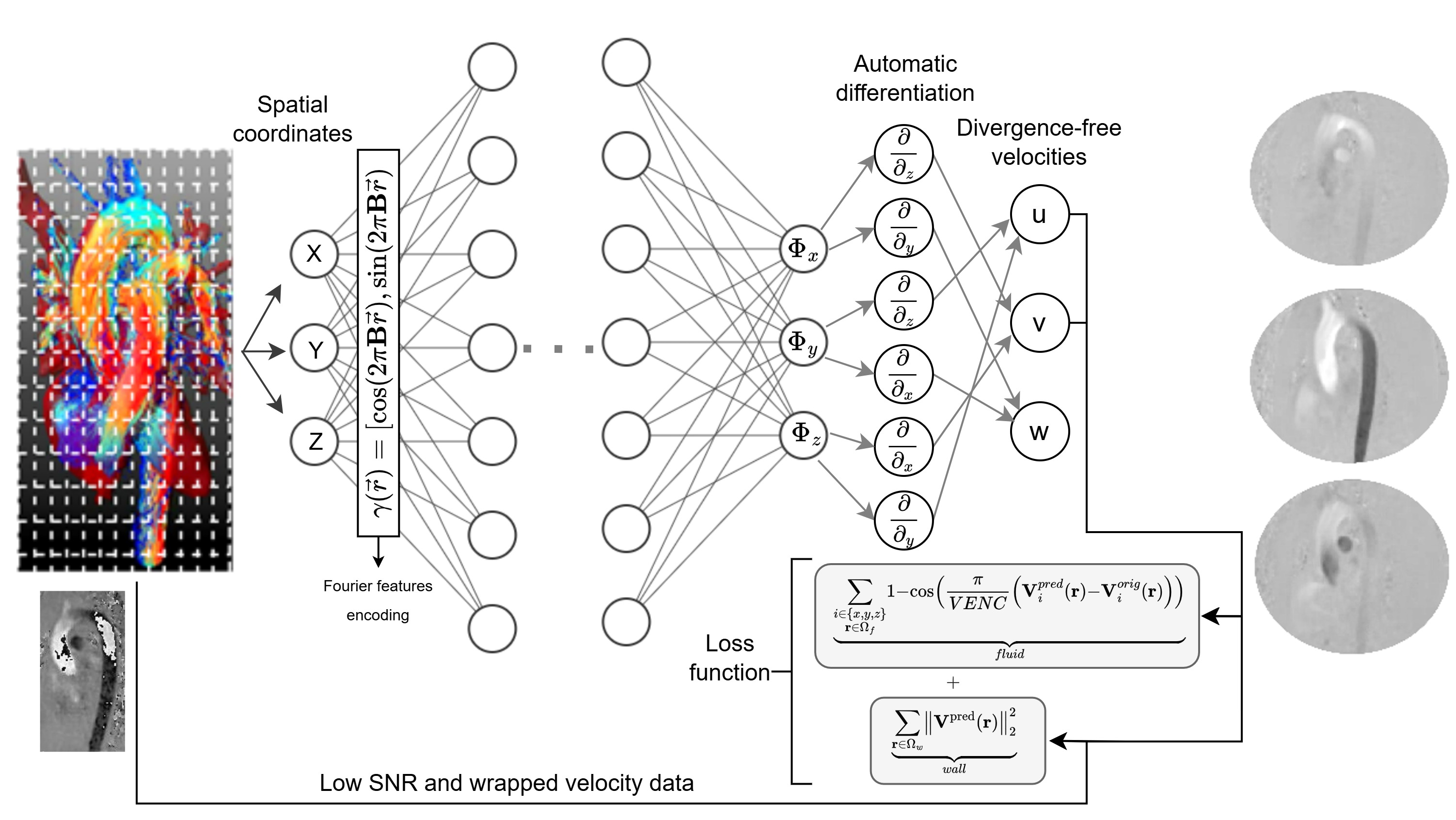}
    \caption{\textbf{Overview of DAF-FlowNet:} Voxel spatial coordinates are encoded using Fourier features and a multilayer perceptron predicts a vector potential  $\mathbf{\Phi}$ whose curl yields divergence-free velocities via automatic differentiation. The network is trained to match the original noisy and wrapped velocities using a cosine-based loss and no-slip boundary condition, enabling simultaneous velocity enhancement and phase unwrapping.}
    \label{fig:Overview}
\end{figure*}

\subsection{Experimental design}

\subsubsection{In silico validation - synthetic 4D Flow MRI}

To evaluate the performance of denoising, divergence minimization, and unwrapping, we used a CFD simulation of a vascular model of the aorta from a 26-year-old healthy male volunteer \cite{jr_ladisa_computational_2011,wilson_vascular_2013}. Using the FEelMRI library\footnote{https://github.com/hmella/feelmri}, we generated synthetic 4D Flow MRI data by solving the Bloch equations and evaluating the MR signal equation using the Finite Elements Method from the CFD simulation \cite{mella2026evaluating}. Trajectories and timings were estimated under realistic hardware constraints for a 1.5T scanner, using scanning parameters obtained from the 2023 4D Flow MRI consensus \cite{bissell_4d_2023}. This physics-based forward model yields synthetic 4D Flow MRI images with realistic MR signal characteristics and acquisition effects, providing a better benchmark for denoising and unwrapping. We generated images with different wrapping and noise levels by varying the VENC during velocity encoding and the percentage of complex Gaussian noise. A detailed list of the parameters is given in \autoref{tab:mri_parameters} and a description of the procedure can be found in the FEelMRI repository.

\begin{table}[!htb]
    \centering
    \caption{Simulation parameters for the generation of synthetic 4D flow MR images. These parameters gave an echo-time of 1.92 ms. Hardware parameters, including maximum gradient amplitude, slew rate, and ADC frequency bandwidth, which dictated sequence timings and image quality, were taken from \cite{Bender2013,dillingerLimitationsEchoPlanar2020}.}
    \label{tab:mri_parameters}
    \begin{tabular}{@{}lc@{}}
        \toprule
        \multicolumn{1}{c}{Imaging parameter} & Value                   \\ \midrule
        VENC (m/s)                    & $2.5$, $1.0$, $0.7$     \\
        Matrix size                   & $120\times 30\times 50$ \\
        Voxel size (mm$^3$)           & $2\times 2\times 2$     \\
        Oversampling factor           & $2$                     \\
        Cardiac phases                & $29$                    \\
        Time spacing (ms)             & $32.0$                  \\
        $T_2^*$ (ms)  \cite{barthProtonNMRRelaxation1997}                  & $254.0$                 \\
        ADC bandwidth (kHz) \cite{Bender2013}           & $128.0$                   \\ 
        Slew-rate (mT/m/s) \cite{dillingerLimitationsEchoPlanar2020}            & $180.0$                   \\
        Max. gradient amplitude (mT/m) \cite{dillingerLimitationsEchoPlanar2020} & $33.0$              \\ \bottomrule
    \end{tabular}
\end{table}

We structured the analysis into three parts: (i) Denoising and divergence minimization, (ii) unwrapping, and (iii) combined enhancement and unwrapping.

For denoising and divergence minimization, we used a high VENC (250 cm/s) to avoid wrapping and tested three noise levels of complex Gaussian noise: 2\%, 4\%, and 6\% of the maximum signal magnitude. We compared DAF-FlowNet against:
\begin{itemize}
    \item \textbf{Divergence Corrector (DivCorr)}: A weighted least-squares error method that adds a corrective velocity to each voxel to minimize divergence \cite{mura_enhancing_2016}.
    \item \textbf{Radial Basis Functions (RBF)}: A projection-based method combining normalized convolution with divergence-free radial basis functions \cite{busch_reconstruction_2013}.
    \item \textbf{Divergence Free Wavelets (DFW)}: A projection-based method that achieves denoising by shrinking divergence-free wavelet and non-divergence-free coefficients \cite{ong_robust_2015}.
    \item \textbf{4DFlowNet}: A deep learning method using a CNN to generate noise-free, super-resolution 4D flow phase images \cite{ferdian_4dflownet_2020}. We used 4DFlowNet's published weights trained on CFD-derived 4D Flow MRI images. 
\end{itemize}
We evaluated our algorithm and the alternatives using three metrics: the normalized root mean square velocity error (VelNRMSE) and the direction error (DE), both computed relative to the reference flow, and the root mean square divergence.  

\begin{align}
    \text { VelNRMSE }&=\left(\max _i\left|\mathbf{v}_i^r\right|\right)^{-1} \sqrt{\frac{1}{N} \sum_i\left|\mathbf{v}_i^r-\mathbf{v}_i^p\right|^2} ,  \label{eq:velnrms} \\
    D E&=\frac{1}{N} \sum_i\left(1-\frac{\mathbf{v}_i^r \cdot \mathbf{v}_i^p}{\left|\mathbf{v}_i^r\right|\left|\mathbf{v}_i^p\right|}\right) , \label{eq:de} \\
    \operatorname{Div} R M S&=\sqrt{\frac{1}{N} \sum_i\left|\operatorname{div}\left(\mathbf{v}_i^{p}\right)\right|^2} \label{eq:divrms}
\end{align}


\noindent
With $\mathbf{v}_i^r$ and $\mathbf{v}_i^p$ the reference and predicted velocity vectors at the \textit{i}th segmented voxel.  To quantify the degradation introduced by simulation and noise, we also report peak velocity to noise ratio (PVNR):

\begin{equation}
    PVNR =20 \log _{10} \frac{1}{\text{VelNRMSE}_{\text{synth}}} \mathrm{dB}
    \label{eq:pvnr}
\end{equation}

\noindent
where $\text{VelNRMSE}_\text{synth}$ is computed between the synthetic and reference velocities.




To assess how different segmentations affect performance, we trained on a dilated mask using a binary structuring element with 18-connectivity. We evaluated the segmentation-dependent methods (DAF-FlowNet, DFW, and RBF) using the dilated mask, and computed the velocity metrics within the original (non-dilated) mask.

For the unwrapping-only evaluation, we simulated phase images with three different VENCs: medium (150 cm/s), low (100 cm/s), and very low (70 cm/s). We implemented three unwrapping methods using a collection of unwrapping tools available in a public repository \cite{dirix_4dflow-unwrap_2024}:

\begin{itemize}
    \item \textbf{Laplacian Unwrapping (Lap4D)}: Unwraps 4D Flow MRI with a 4D single-step laplacian algorithm \cite{loecher_phase_2016}.
    \item \textbf{Non-Continuous Path with Reliability Sorting (NPRS)}: Employs a path-following scheme in the unwrapping process, incorporating a reliability sorting mechanism \cite{abdul-rahman_fast_2005}.
    \item \textbf{Graph Cuts 3D Unwrapping (GC3D)}: Applies a graph-based method tailored for 3D data to unwrap phase discontinuities \cite{bioucas-dias_phase_2007}.
\end{itemize}

We assessed the performance of the unwrapping process by measuring the percentage of remaining wrapped voxels. A voxel was considered wrapped if the deviation $| \hat{\phi} - \phi | $ was greater than $\pi$, where $\hat{\phi}$ is the obtained phase after the unwrapping algorithm, and $\phi$ is the simulated phase for high VENC (250 cm/s).

For joint phase unwrapping and velocity enhancement, we used synthetic data with $\mathrm{VENC}=100$~cm/s and the same complex Gaussian noise levels (2\%, 4\%, and 6\%). We compared DAF-FlowNet against the sequential pipeline combining the best-performing alternatives for unwrapping and enhancement.

\subsubsection{In-vivo 4D Flow MRI}

To evaluate clinical applicability, we tested our approach on 4D Flow MRI data previously obtained from 10 patients with hypertrophic cardiomyopathy (HCM). A previous study was approved by the respective ethics committee \cite{iwata_measurement_2024}, and the data was used for a secondary purpose in this study. All subjects provided written informed consent prior to the examinations. Among them, 5 patients presented with left ventricular outflow tract obstruction. We refer to patients with and without obstruction as HOCM and HNCM, respectively.

Datasets were acquired with a 3.0T Achieva MRI scanner (Philips Healthcare, Best, the Netherlands). Imaging parameters were as follows: TR/TE 4.0/2.7 ms; flip angle 11°; acquired voxel resolution of 1.70 × 1.70 × 2.00 mm with zero-fill interpolation in the slice-direction; multivelocity encoding (VENC) acquisition of 50–150–450 cm/s; temporal resolution of 40 ms; 15–21 cardiac phases depending on heart rate; prospective triggering; k-t principal component analysis (PCA) acceleration factor of 5–7; free breathing acquisition with an abdominal belt for restricting motion; and acquisition time of 10–15 min \cite{binter_bayesian_2013,takahashi2021four}. For this study, we processed only acquisitions with VENC 150 cm/s, as 50 cm/s exhibited severe wrapping and 450 cm/s had very low VNR.

To assess quality in  both original and predicted velocities, we performed a mass conservation analysis, as recommended in the 2023 4D Flow MRI consensus \cite{bissell_4d_2023}, using flow comparisons between:
\begin{itemize}
    \item Two ascending aorta planes, between the sinuses of Valsalva and the first branching vessel (AAoP and AAoD).
    \item The right and left pulmonary arteries (RPA and LPA) vs. the main pulmonary artery (PA).
\end{itemize}

We report absolute bias (L/min) and relative bias (\%) between planes for each scenario.

We compared DAF-FlowNet against the best-performing alternatives on synthetic data. For velocity enhancement, we compared against DFW; however, thresholding the divergence-free and non-divergence-free components using \textit{SureShrink} was overly conservative, leaving the in-vivo velocities practically unchanged. Therefore, we performed manual thresholding using the L-curve criterion \cite{hansen1993use} as suggested in the original paper \cite{ong_robust_2015}. For joint velocity enhancement and phase unwrapping, we compared against unwrapping with GC3D followed by denoising and divergence minimization with DFW.


\subsection{Implementation details}

Optimization was performed on an NVIDIA A100 Tensor Core GPU with 40 GB of memory using the Adam optimizer. The number of coordinates sampled per backpropagation step was set to the minimum of 50,000 and the total number of segmented lumen and wall coordinates, in order to maximize GPU memory utilization. The complete implementation is publicly available at \href{https://github.com/JavierBZ/DAF-FlowNet}{https://github.com/JavierBZ/DAF-FlowNet}.

\section{Results}

\subsection{In-silico validation}

\subsubsection{Denoising and divergence minimization (no wrapping)}

\begin{figure*}[!ht]
    \centering
    \includegraphics[width=\textwidth,]{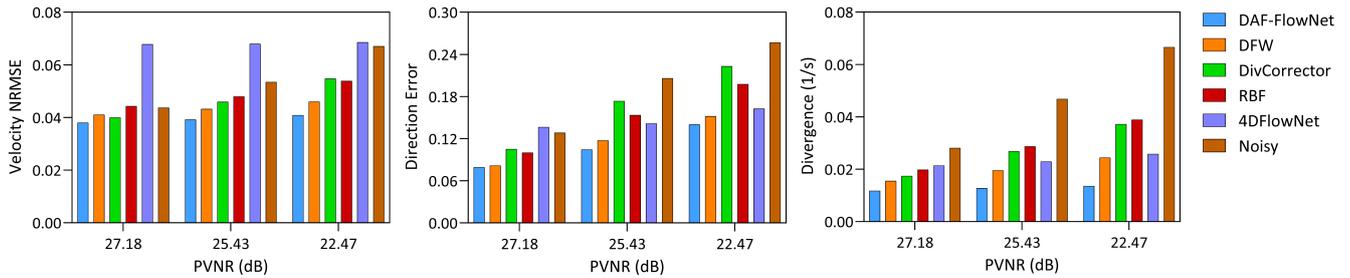}
    \caption{Synthetic (no-wrap) denoising and divergence minimization benchmark ($\text{VENC}=250$ cm/s). Quantitative comparison across noise levels using velocity NRMSE, direction error, and RMS divergence for DAF-FlowNet and divergence-free/physics-based alternatives.}
    \label{fig:SimBars}
\end{figure*}

\begin{figure*}[!ht]
    \centering
    \includegraphics[width=\textwidth,]{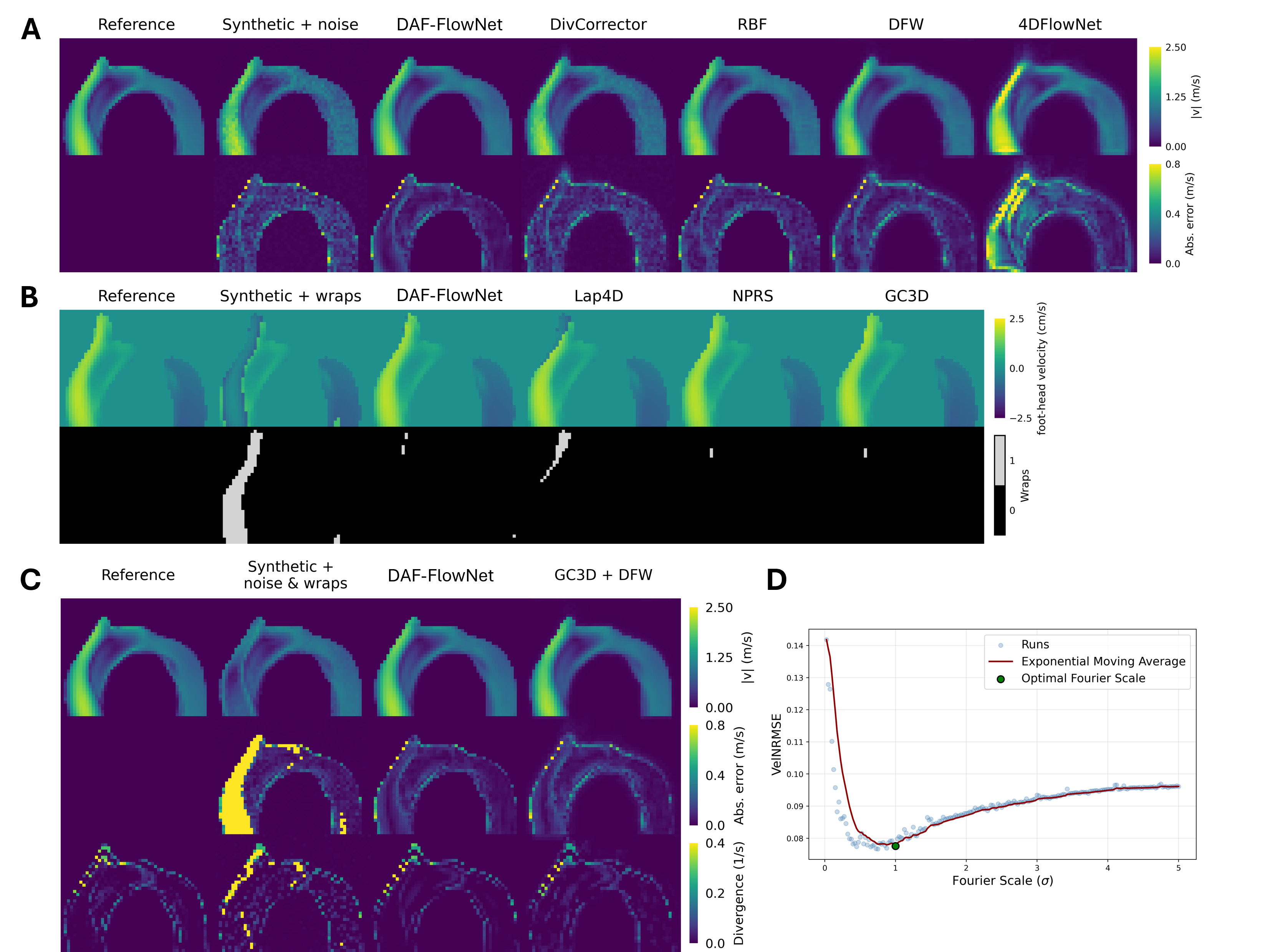}
    \caption{Synthetic experiments: (A) Velocity magnitude maps and absolute error maps for reference, noisy input, and denoising/divergence minimization outputs. (B) Foot–head velocity component for a simulation without wraps, wrapped velocity ($\text{VENC} = 100$ cm/s), and unwrapped outputs; bottom row indicates wrapped-voxel locations. (C) Combined velocity enhancement and unwrapping. (D) Fourier scale ($\sigma$) sweep analysis.}
    \label{fig:Sim}
\end{figure*}

DAF-FlowNet outperformed the competing approaches in terms of velocity NRMSE, direction error, and divergence (\autoref{fig:SimBars}). Qualitatively, DAF-FlowNet suppressed noise while preserving the overall flow patterns observed in the reference (\autoref{fig:Sim}A). The error maps indicate strong noise suppression within the lumen, while residual errors are more pronounced near the vessel boundary, a trend also observed across methods.

\begin{table}[!ht]
\centering
\small
\setlength{\tabcolsep}{4pt}
\caption{Sensitivity to segmentation: effect of mask dilation on denoising metrics. Results are reported as original-mask $\rightarrow$ dilated-mask.}
\label{tab:mask_dilation}
\begin{tabular}{llccc}
\toprule
PVNR & Metric & DAF-FlowNet & RBF & DFW \\
\midrule
\multirow{3}{*}{27.18 dB}
 & VelNRMSE                 & 0.038$\rightarrow$0.038 & 0.044$\rightarrow$0.045 & 0.041$\rightarrow$0.042 \\
 & DE                       & 0.079$\rightarrow$0.079 & 0.100$\rightarrow$0.101 & 0.082$\rightarrow$0.108 \\
 & Div. (1/s) & 0.012$\rightarrow$0.010 & 0.020$\rightarrow$0.013 & 0.016$\rightarrow$0.022 \\
\midrule
\multirow{3}{*}{25.43 dB}
 & VelNRMSE                 & 0.039$\rightarrow$0.039 & 0.048$\rightarrow$0.047 & 0.043$\rightarrow$0.047 \\
 & DE                       & 0.105$\rightarrow$0.101 & 0.154$\rightarrow$0.143 & 0.117$\rightarrow$0.175 \\
 & Div. (1/s) & 0.013$\rightarrow$0.011 & 0.029$\rightarrow$0.019 & 0.020$\rightarrow$0.035 \\
\midrule
\multirow{3}{*}{23.47 dB}
 & VelNRMSE                 & 0.041$\rightarrow$0.041 & 0.054$\rightarrow$0.050 & 0.046$\rightarrow$0.055 \\
 & DE                       & 0.140$\rightarrow$0.136 & 0.198$\rightarrow$0.182 & 0.152$\rightarrow$0.223 \\
 & Div. (1/s) & 0.014$\rightarrow$0.011 & 0.039$\rightarrow$0.026 & 0.024$\rightarrow$0.048 \\
\bottomrule
\end{tabular}
\end{table}

Across all noise levels, DAF-FlowNet was largely insensitive to mask dilation, with negligible changes in VelNRMSE, DE and divergence (\autoref{tab:mask_dilation}). RBF showed slightly larger variations in VelNRMSE, while improving DE and Div${\mathrm{RMS}}$ after dilation. In contrast, DFW was sensitive to dilation, with consistent degradations particularly in DE and Div${\mathrm{RMS}}$ that became more pronounced as noise increased.

\subsubsection{Velocity unwrapping}

For medium VENC (150~cm/s) and low VENC (100~cm/s), DAF-FlowNet produced the lowest residual wrapping with and without noise (\autoref{tab:unwrap}). For very low VENC (70~cm/s), DAF-FlowNet left substantially more residual wraps ($\approx$37--39\%), whereas NPRS/GC3D yielded the lowest residual wrapping ($\approx$7--10\%). Overall, DAF-FlowNet is the most effective in the medium-to-low VENC regime (\autoref{fig:Sim}B).

\begin{table}[!ht]
\centering
\small
\caption{Remaining wrapped voxels (\%) for each method with different VENC values with and without noise ($\text{PVNR}\approx 25 dB$). Bold indicates best (lowest) value per column. As reference, the maximum velocity of the CFD simulation is 207 cm/s.}
\begin{tabular}{c|ccc|ccc}
\toprule
\multirow{3}{*}{\raisebox{-1.6ex}{Method}} 
    & \multicolumn{6}{c}{VENC (cm/s)} \\
\cmidrule(lr){2-7}
    & \multicolumn{3}{c|}{Only wraps} 
    & \multicolumn{3}{c}{Wraps + complex noise} \\
\cmidrule(lr){2-7}
    & 150 & 100 & 70 
    & 150 & 100 & 70 \\
\midrule
DAF-FlowNet 
    & \textbf{0.18} & \textbf{5.17} & 38.66
    & \textbf{0.18} & \textbf{4.92} & 37.39 \\
Lap4D 
    & 0.65 & 11.14 & 31.97
    & 0.76 & 12.18 & 37.81 \\
NPRS 
    & 1.12 & 7.88 & 7.16
    & 1.17 & 7.16 & 10.17 \\
GC3D 
    & 0.83 & 6.31 & \textbf{6.83}
    & 0.88 & 6.59 & \textbf{8.98} \\
\bottomrule
\end{tabular}
\label{tab:unwrap}
\end{table}





\subsubsection{Combined enhancement and unwrapping}

At $\mathrm{VENC}=100$~cm/s across all noise levels, DAF-FlowNet achieved lower VelNRMSE, lower divergence, and fewer wrapped voxels than the best sequential baseline (GC3D + DFW) in all three cases. DE was also reduced at medium noise and high noise, whereas at the lowest noise GC3D + DFW and DAF-FlowNet achieved the same DE.

\begin{table}[t]
\centering
\small
\setlength{\tabcolsep}{5pt}
\caption{Combined analysis (enhancement + unwrapping) at $\mathrm{VENC}=100$~cm/s. Results compare DAF-FlowNet against the best sequential baseline (GC3D + DFW). PVNR was computed on the fluid region without wraps.}
\label{tab:combined_unwrap_enhance}
\begin{tabular}{llcc}
\toprule
PVNR& Metric & DAF-FlowNet & GC3D + DFW \\
\midrule
\multirow{4}{*}{24.57 dB}
 & VelNRMSE                 & $\mathbf{0.050}$ & $0.057$ \\
 & DE                     & $\mathbf{0.060}$ & $\mathbf{0.060}$ \\
 & Divergence (1/s)       & $\mathbf{0.013}$ & $0.015$ \\
 & Wrapped voxels (\%)    & $\mathbf{5.280}$ & $6.303$ \\
\midrule
\multirow{4}{*}{24.32 dB}
 & VelNRMSE                 & $\mathbf{0.050}$ & $0.059$ \\
 & DE                     & $\mathbf{ 0.067}$ & $0.075$ \\
 & Divergence (1/s)       & $\mathbf{0.013}$ & $0.017$ \\
 & Wrapped voxels (\%)    & $\mathbf{4.922}$ & $6.028$ \\
\midrule
\multirow{4}{*}{23.93 dB}
 & VelNRMSE                 & $\mathbf{0.050}$ & $0.057$ \\
 & DE                     & $\mathbf{0.081}$ & $0.091$ \\
 & Divergence (1/s)       & $\mathbf{0.014}$ & $0.018$ \\
 & Wrapped voxels (\%)    & $\mathbf{4.933}$ & $6.904$ \\
\bottomrule
\end{tabular}
\end{table}



\subsubsection{Hyperparameter tuning and Fourier scale}
\begin{figure}[!ht]
    \centering
    \includegraphics[width=0.95\columnwidth,]{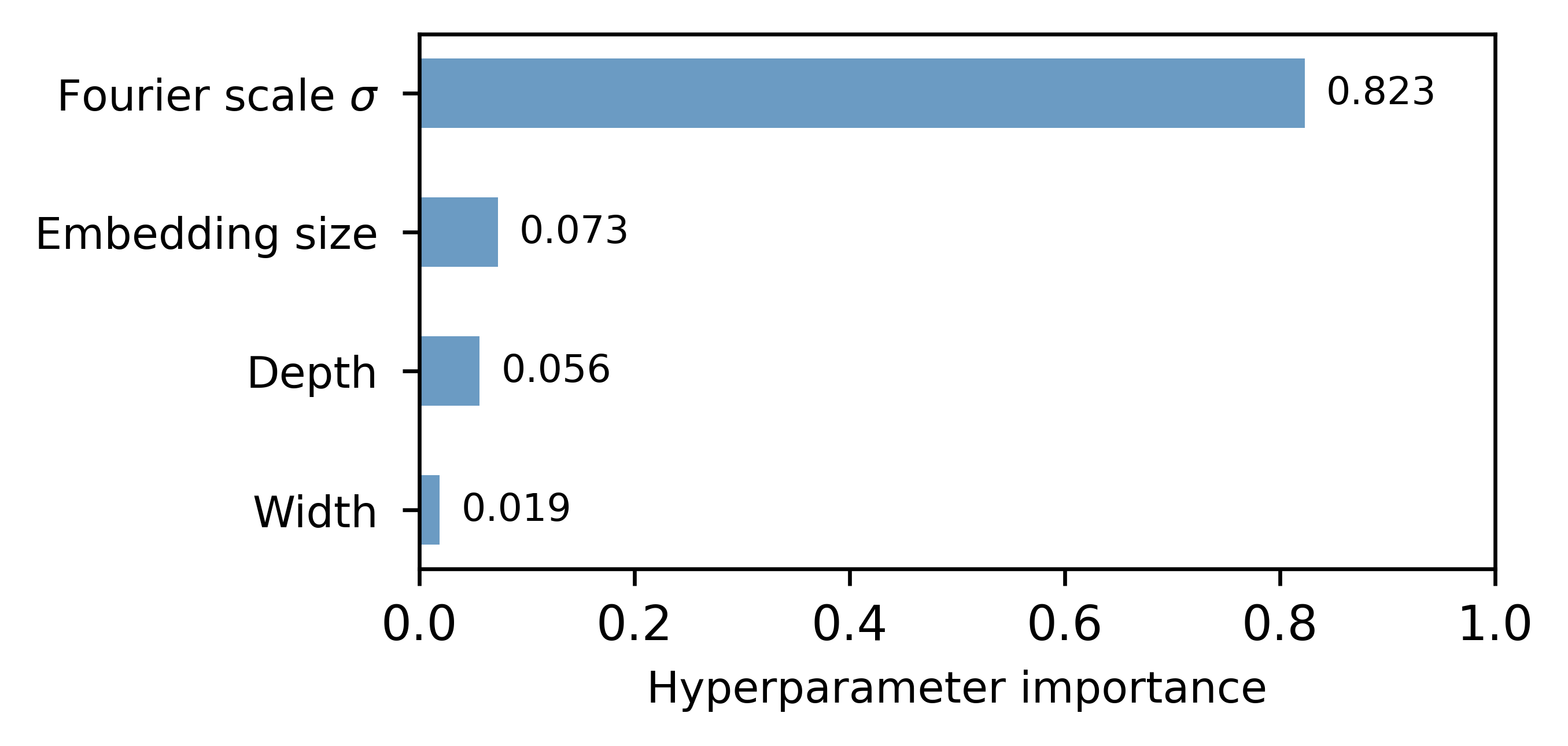}
    \caption{Hyperparameter sensitivity from the grid search. Relative influence of the evaluated architectural parameters on validation error, highlighting that the Fourier scale $\sigma$ dominates performance compared with embedding size, depth, and width. }
    \label{fig:Grid_searchs}
\end{figure}

To understand the sensitivity of the previous results to training configuration, we examined the relative importance of each architecture hyperparameter, based on the lowest VelNRMSE on the intermediate noise level (PVNR = 25.43 dB). Our grid-search analysis indicated that reconstruction accuracy is primarily governed by the Fourier scale $\sigma$ with a parameter importance of 0.823 (\autoref{fig:Grid_searchs}). For the sweep over $\sigma$, the VelNRMSE curve exhibited an optimum near $\sigma=1$. Small $\sigma$ underfit high-frequency structures and large $\sigma$ introduced noise into the reconstructions (\autoref{fig:Sim}D). 

\begin{figure*}[!ht]
    \centering
    \includegraphics[width=0.99\textwidth,]{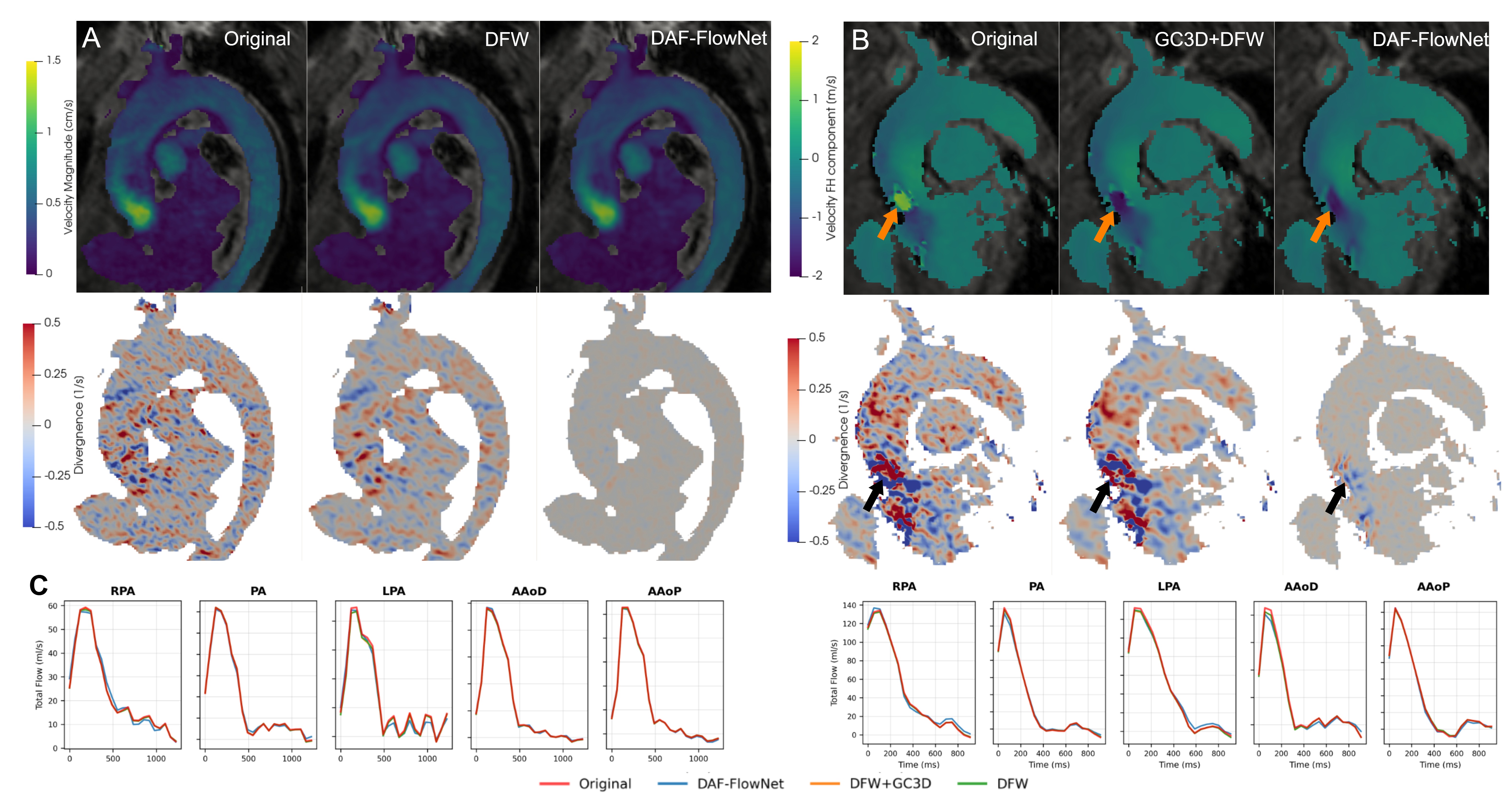}
    \caption{In-vivo validation. (A) HNCM example: velocity magnitude (top row) and divergence maps (middle row) for Original, DFW, and DAF-FlowNet; arrows highlight fine structure smoothed by DFW and preserved by DAF-FlowNet. Bottom: absolute differences vs. Original. (B) HOCM example: velocity magnitudes for Original, GC3D + DFW, and DAF-FlowNet; arrows indicate wrapped regions corrected by GC3D and DAF-FlowNet. (C) Flow waveforms at pulmonary and aortic planes; top row corresponds to (A) and bottom row to (B).}
    \label{fig:InVivo}
\end{figure*}

\begin{table*}[!ht]
\centering
\small
\caption{In-vivo flow quantification and mass-conservation analysis for the HNCM (n=5) and HOCM (n=5) cohorts. The upper section reports plane-based mean flow rates (L/min; mean ± SD) for the original acquisition and each post-processing method. The lower section reports mass-conservation error as absolute bias (L/min) and relative bias (\%) using paired aortic and pulmonary planes.}
\label{tab:invivo_analysis}
\setlength{\tabcolsep}{2.8pt}
\begin{tabular}{@{}lcccccccc@{}}
\toprule
\multirow{2}{*}[-1.5ex]{\textbf{Flow (L/min)}} &
\multicolumn{4}{c}{\textbf{HNCM}} &
\multicolumn{4}{c}{\textbf{HOCM}} \\
\cmidrule(lr){2-5} \cmidrule(lr){6-9}
&
\textbf{Original} &
\textbf{DAF-FlowNet} &
\textbf{GC3D+DFW} &
\textbf{DFW} &
\textbf{Original} &
\textbf{DAF-FlowNet} &
\textbf{GC3D+DFW} &
\textbf{DFW} \\
\midrule
RPA  & $2.42 \pm 1.07$ & $2.37 \pm 1.10$ & $2.39 \pm 1.06$ & $2.39 \pm 1.06$
     & $2.65 \pm 0.69$ & $2.57 \pm 0.71$ & $2.57 \pm 0.69$ & $2.61 \pm 0.65$ \\
PA   & $3.68 \pm 1.66$ & $3.64 \pm 1.60$ & $3.66 \pm 1.65$ & $3.65 \pm 1.65$
     & $4.26 \pm 1.04$ & $4.29 \pm 1.04$ & $4.23 \pm 1.00$ & $4.22 \pm 1.03$ \\
LPA  & $1.27 \pm 0.58$ & $1.28 \pm 0.64$ & $1.23 \pm 0.59$ & $1.23 \pm 0.59$
     & $1.63 \pm 0.58$ & $1.68 \pm 0.53$ & $1.56 \pm 0.57$ & $1.56 \pm 0.57$ \\
AAoD & $3.92 \pm 0.86$ & $3.84 \pm 0.86$ & $3.89 \pm 0.85$ & $3.89 \pm 0.85$
     & $4.72 \pm 1.41$ & $4.51 \pm 1.41$ & $4.60 \pm 1.37$ & $4.63 \pm 1.42$ \\
AAoP & $3.89 \pm 1.03$ & $3.75 \pm 0.91$ & $3.83 \pm 1.01$ & $3.83 \pm 1.01$
     & $4.72 \pm 1.25$ & $4.54 \pm 1.29$ & $4.74 \pm 1.23$ & $4.73 \pm 1.23$ \\
\midrule
\textbf{Mass conservation} &
\textbf{Original} & \textbf{DAF-FlowNet} & \textbf{GC3D+DFW} & \textbf{DFW} &
\textbf{Original} & \textbf{DAF-FlowNet} & \textbf{GC3D+DFW} & \textbf{DFW} \\
\midrule
AAoP-AAoD (L/min) &
$0.24 \pm 0.14$ & $\mathbf{0.12 \pm 0.10}$ & $0.25 \pm 0.09$ & $0.25 \pm 0.09$ &
$0.24 \pm 0.20$ & $\mathbf{0.20 \pm 0.20}$ & $0.23 \pm 0.28$ & $0.27 \pm 0.27$ \\
AAoP-AAoD (\%)    &
$6.39 \pm 3.57$ & $\mathbf{3.62 \pm 4.00}$ & $6.85 \pm 3.77$ & $6.92 \pm 3.62$ &
$5.73 \pm 5.43$ & $\mathbf{4.97 \pm 5.87}$ & $5.78 \pm 8.02$ & $6.47 \pm 7.70$ \\
PA-(RPA+LPA) (L/min) &
$0.18 \pm 0.09$ & $\mathbf{0.07 \pm 0.06}$ & $0.21 \pm 0.15$ & $0.21 \pm 0.15$ &
$0.38 \pm 0.18$ & $\mathbf{0.24 \pm 0.28}$ & $0.31 \pm 0.19$ & $0.35 \pm 0.11$ \\
PA-(RPA+LPA) (\%)   &
$4.91 \pm 1.82$ & $\mathbf{2.00 \pm 1.47}$ & $5.44 \pm 3.20$ & $5.40 \pm 3.24$ &
$8.79 \pm 2.79$ & $\mathbf{5.07 \pm 5.89}$ & $7.00 \pm 3.88$ & $8.51 \pm 1.80$ \\
\bottomrule
\end{tabular}
\end{table*}

\subsection{In-vivo validation: qualitative assessment and mass conservation analysis}

DAF-FlowNet and DFW showed clear denoising while preserving the original velocity features. However, only DAF-FlowNet achieved substantial divergence reduction compared to the original acquisitions (\autoref{fig:InVivo}A-B). In the HOCM example with aliasing, both the sequential baseline (DFW+GC3D) and DAF-FlowNet corrected the wrapped regions (\autoref{fig:InVivo}B). Flow waveforms at the pulmonary and aortic analysis planes remained consistent with the original measurements for the representative HNCM and HOCM subjects shown in \autoref{fig:InVivo}C, with small plane-wise deviations (mean relative differences $<$5\% in both groups; \autoref{tab:invivo_analysis}).

Mass conservation improved with DAF-FlowNet relative to the original data and the alternative enhancement and unwrapping methods. For HNCM, AAo inter-plane bias decreased from \(0.24\pm0.14\) to \(0.12\pm0.10\)~L/min (relative bias \(6.39\pm3.57\%\) to \(3.62\pm4.00\%\)), and pulmonary bias decreased from \(0.18\pm0.09\) to \(0.07\pm0.06\)~L/min (relative bias \(4.91\pm1.82\%\) to \(2.00\pm1.47\%\)). For HOCM, pulmonary bias decreased from \(0.38\pm0.18\) to \(0.24\pm0.28\)~L/min (relative bias \(8.79\pm2.79\%\) to \(5.07\pm5.89\%\)) (\autoref{tab:invivo_analysis}).

\subsection{Execution times} 
\label{execution_times}

\begin{table}[t]
\centering
\small
\setlength{\tabcolsep}{5pt}
\caption{Execution times per timeframe (s) and total runtime (HH:MM:SS). Simulation has 29 timeframes (TF); in-vivo has median 18 timeframes.}
\label{tab:exec_times}
\begin{tabular}{llcc}
\toprule
Dataset & Method & Per TF (s) & Total\\
\midrule
\multirow{8}{*}{Synthetic} 
& DAF-FlowNet & 94.00 & 00:45:26 \\
& DivCorr    & 62.00 & 00:29:58 \\
& 4DFlowNet  & 1.03  & 00:00:30 \\
& RBF        & 5.89  & 00:02:51 \\
& DFW        & 0.23  & 00:00:07 \\
\cmidrule(lr){2-4}
& Lap4D & 0.140 & 00:00:04 \\
& NPRS  & 0.502 & 00:00:15 \\
& GC3D  & 0.963 & 00:00:28 \\
\midrule
\multirow{3}{*}{In-vivo} 
& DAF-FlowNet & 475.32 & 02:22:36 \\
& DFW        & 5.02   & 00:01:30 \\
& GC3D & 41.23 & 00:12:22 \\
\bottomrule
\end{tabular}
\end{table}

Execution times (\autoref{tab:exec_times}) are reported per timeframe (s) and as totals across all timeframes (HH:MM:SS) to account for the differing number of timeframes. Overall, DAF-FlowNet is the most computationally demanding, whereas DFW for velocity enhancement and Lap4D for unwrapping are substantially faster.

\section{Discussion}\label{sec:discussion}

The synthetic validation demonstrates that DAF-FlowNet's parameterization of velocities as the curl of a vector potential offers a practical advantage over existing velocity-enhancement methods, yielding improved divergence, VelNRMSE, and DE. Including the vector potential allows the network to predict the best-fitting divergence-free velocity field for the data, without balancing physics regularization, thereby improving generalization. Combined with the cosine data-consistency loss, this architectural choice allows the model to address phase unwrapping and denoising as a single, unified optimization problem.


The results highlight that DAF-FlowNet is particularly robust to moderate segmentation inaccuracies. Traditional projection-based methods, such as DFW, exhibit performance degradation when masks are dilated, likely due to the propagation of zero-velocity background noise into the lumen. DAF-FlowNet maintains stable performance metrics. This suggests that the coordinate-based neural representation is better at learning the underlying flow manifold even when the input domain is imperfectly defined.
Furthermore, the hyperparameter analysis confirms that the Fourier scale ($\sigma$) is the critical parameter for balancing denoising and feature preservation. An optimal value near 1 for the synthetic dataset allows the network to capture high-frequency flow features without fitting the high-frequency components of noise and simulation artifacts. With appropriate scaling using $\tau$ (\autoref{eq:scaling_factor}), the Fourier feature encoding can be adapted to a new domain.

The in-vivo validation in HCM patients underscores the clinical utility of DAF-FlowNet, particularly in its ability to preserve flow waveforms while substantially reducing divergence. On the other hand, DFW can suppress noise but yields a smaller reduction in divergence, resulting in less agreement for mass conservation. This is clinically relevant because it directly improves the internal consistency between paired aortic and pulmonary measurements in routine flow quantification, thereby improving quality-control mass-conservation analysis recommended for 4D Flow MRI workflows \cite{bissell_4d_2023}. Moreover, the ability to correct aliased regions in obstructive cases within the same formulation suggests a practical route to leverage lower VENCs (higher VNR) without relying on a separate unwrapping step. Nevertheless, broader multi-center validation remains necessary.

Recent preprints have also addressed related aspects of 4D Flow post-processing. SMURF \cite{hans2025smurf} demonstrates the value of probabilistic coordinate-based neural representations for joint segmentation and reconstruction, but without validation against alternative denoising methods. VAST \cite{singh2026vast} addresses the coupled unwrapping-denoising problem for intracranial applications using continuity constraints and POD-based denoising,  but without benchmarking against alternative unwrapping or denoising methods. In contrast, DAF-FlowNet was validated against multiple competing methods for both denoising and unwrapping.




\subsection*{Limitations and future directions}

DAF-FlowNet is less effective in very low-VENC setups as the number of local optima increases with multiple wraps. In such cases, more robust unwrapping approaches (e.g., NPRS or GC3D) could serve as complementary tools, while DAF-FlowNet focuses on velocity denoising and divergence minimization. 

DAF-FlowNet leverages the cosine data-consistency formulation in \cite{carrillo_optimal_2019} to handle wrapped phase. Nevertheless, that formulation was originally developed for dual or multi-VENC settings, where additional measurements ensure global optima. A direct next step is to extend DAF-FlowNet to explicit multi-VENC reconstruction so that the extra encodings can improve both enhancement and unwrapping.


Domain transfer is handled via the proposed scaling factor ($\tau$; \autoref{eq:scaling_factor}), which adapts the Fourier feature representation of normalized coordinates to changes in image size and spatial resolution. While this provides a simple and interpretable mechanism, feature frequencies may shift across domains, and the optimal Fourier scale may differ even after scaling. Methods such as spatially adaptive progressive encoding \cite{hertz2021sape} or frequency shifting \cite{kania2024fresh} could improve the tuning of spectral bias.

DAF-FlowNet’s coordinate-based representation also enables predictions at arbitrary spatial resolutions, which may facilitate super-resolution 4D Flow MRI. Prior work has shown that sinusoidal neural representations (SIRENs) can support denoising and spatiotemporal super-resolution of 4D Flow velocity fields \cite{saitta_implicit_2024}. In this context, the divergence-free parameterization provides an implicit physical structure that could enhance their generalization.

Finally, training time remains a practical limitation relative to alternative techniques (\autoref{execution_times}). Parallel execution across cardiac phases, including training multiple networks concurrently on a single GPU and distributing them across multiple GPUs, could substantially reduce training time and improve feasibility for larger cohorts. In addition, using learned weight initializations through meta-learning strategies may further improve both training efficiency and reconstruction quality \cite{tancik2021learned}.



\section{Conclusion}

DAF-FlowNet demonstrates that parameterizing velocities as the curl of a vector potential and enforcing mass conservation by architecture rather than a penalty term yields measurable accuracy gains over existing methods while removing a key source of per-dataset tuning. The cosine data-consistency loss extends this framework to handle phase wrapping within the same optimization, eliminating the need for a separate unwrapping step. In-vivo results in HCM patients confirm that these properties carry over to clinical data, producing consistent improvements in aortic and pulmonary mass-conservation analyses recommended by the 4D Flow MRI consensus guidelines. These findings position physics-constrained coordinate-based representations as a practical foundation for cardiovascular 4D Flow MRI post-processing, with natural extensions toward multi-VENC acquisitions and broader anatomical validation.

\section*{Acknowledgment}
J.S. thanks the Department of Medical Imaging and Radiation Sciences at Monash University.


\bibliography{4Dflow}

@article{barthProtonNMRRelaxation1997,
  title = {Proton {{NMR}} Relaxation Times of Human Blood Samples at 1.5 {{T}} and Implications for Functional {{MRI}}},
  author = {Barth, M. and Moser, E.},
  year = {1997},
  month = jul,
  journal = {Cellular and Molecular Biology (Noisy-Le-Grand, France)},
  volume = {43},
  number = {5},
  pages = {783--791},
  issn = {0145-5680},
  langid = {english},
  pmid = {9298600}
}

@misc{wandb,
title = {Experiment Tracking with Weights and Biases},
year = {2020},
note = {Software available from wandb.com},
url={https://www.wandb.com/},
author = {Biewald, Lukas},
}

@article{Bender2013,
  title = {The Importance of K-Space Trajectory on off-Resonance Artifact in Segmented Echo-Planar Imaging},
  author = {Bender, Jacob A. and Ahmad, Rizwan and Simonetti, Orlando P.},
  year = {2013},
  month = mar,
  journal = {Concepts in Magnetic Resonance Part A: Bridging Education and Research},
  volume = {42 A},
  number = {2},
  pages = {23--31},
  issn = {15466086},
  doi = {10.1002/cmr.a.21255}
}

@article{dillingerLimitationsEchoPlanar2020,
  title = {On the Limitations of Echo Planar {{4D}} Flow {{MRI}}},
  author = {Dillinger, Hannes and Walheim, Jonas and Kozerke, Sebastian},
  year = {2020},
  journal = {Magnetic Resonance in Medicine},
  volume = {84},
  number = {4},
  pages = {1806--1816},
  issn = {1522-2594},
  doi = {10.1002/mrm.28236},
  langid = {english}
}

@article{markl_4d_2012,
  title={4D flow MRI},
  author={Markl, Michael and Frydrychowicz, Alex and Kozerke, Sebastian and Hope, Mike and Wieben, Oliver},
  journal={Journal of Magnetic Resonance Imaging},
  volume={36},
  number={5},
  pages={1015--1036},
  year={2012},
  publisher={Wiley Online Library}
}

@article{sotelo_fully_2022,
  title={Fully three-dimensional hemodynamic characterization of altered blood flow in bicuspid aortic valve patients with respect to aortic dilatation: a finite element approach},
  author={Sotelo, Julio and Franco, Pamela and Guala, Andrea and Dux-Santoy, Lydia and Ruiz-Mu{\~n}oz, Aroa and Evangelista, Arturo and Mella, Hernan and Mura, Joaqu{\'\i}n and Hurtado, Daniel E and Rodr{\'\i}guez-Palomares, Jos{\'e} F and others},
  journal={Frontiers in Cardiovascular Medicine},
  volume={9},
  pages={885338},
  year={2022},
  publisher={Frontiers Media SA}
}

@article{jiang_quantifying_2015,
  title={Quantifying errors in flow measurement using phase contrast magnetic resonance imaging: comparison of several boundary detection methods},
  author={Jiang, Jing and Kokeny, Paul and Ying, Wang and Magnano, Chris and Zivadinov, Robert and Haacke, E Mark},
  journal={Magnetic resonance imaging},
  volume={33},
  number={2},
  pages={185--193},
  year={2015},
  publisher={Elsevier}
}

@article{ko_novel_2019,
  title={Novel and facile criterion to assess the accuracy of WSS estimation by 4D flow MRI},
  author={Ko, Seungbin and Yang, Byungkuen and Cho, Jee-Hyun and Lee, Jeesoo and Song, Simon},
  journal={Medical image analysis},
  volume={53},
  pages={95--103},
  year={2019},
  publisher={Elsevier}
}

@article{petersson_assessment_2012,
  title={Assessment of the accuracy of MRI wall shear stress estimation using numerical simulations},
  author={Petersson, Sven and Dyverfeldt, Petter and Ebbers, Tino},
  journal={Journal of Magnetic Resonance Imaging},
  volume={36},
  number={1},
  pages={128--138},
  year={2012},
  publisher={Wiley Online Library}
}

@article{marlevi_estimation_2019,
  title={Estimation of cardiovascular relative pressure using virtual work-energy},
  author={Marlevi, David and Ruijsink, Bram and Balmus, Maximilian and Dillon-Murphy, Desmond and Fovargue, Daniel and Pushparajah, Kuberan and Bertoglio, Crist{\'o}bal and Colarieti-Tosti, Massimiliano and Larsson, Matilda and Lamata, Pablo and others},
  journal={Scientific reports},
  volume={9},
  number={1},
  pages={1375},
  year={2019},
  publisher={Nature Publishing Group UK London}
}

@article{bostan_improved_2014,
  title={Improved variational denoising of flow fields with application to phase-contrast MRI data},
  author={Bostan, Emrah and Lefkimmiatis, Stamatios and Vardoulis, Orestis and Stergiopulos, Nikolaos and Unser, Michael},
  journal={IEEE Signal Processing Letters},
  volume={22},
  number={6},
  pages={762--766},
  year={2014},
  publisher={IEEE}
}

@article{mura_enhancing_2016,
  title={Enhancing the velocity data from 4D flow MR images by reducing its divergence},
  author={Mura, Joaqu{\'\i}n and Pino, A Mat{\'\i}as and Sotelo, Julio and Valverde, Israel and Tejos, Cristian and Andia, Marcelo E and Irarrazaval, Pablo and Uribe, Sergio},
  journal={IEEE transactions on medical imaging},
  volume={35},
  number={10},
  pages={2353--2364},
  year={2016},
  publisher={IEEE}
}

@article{zhang_divergence-free_2021,
  title={Divergence-free constrained phase unwrapping and denoising for 4D flow MRI using weighted least-squares},
  author={Zhang, Jiacheng and Rothenberger, Sean M and Brindise, Melissa C and Scott, Michael B and Berhane, Haben and Baraboo, Justin J and Markl, Michael and Rayz, Vitaliy L and Vlachos, Pavlos P},
  journal={IEEE transactions on medical imaging},
  volume={40},
  number={12},
  pages={3389--3399},
  year={2021},
  publisher={IEEE}
}

@article{deriaz_divergence-free_2006,
  title={Divergence-free and curl-free wavelets in two dimensions and three dimensions: application to turbulent flows},
  author={Deriaz, Erwan and Perrier, Val{\'e}rie},
  journal={Journal of Turbulence},
  number={7},
  pages={N3},
  year={2006},
  publisher={Taylor \& Francis}
}

@article{busch_reconstruction_2013,
  title={Reconstruction of divergence-free velocity fields from cine 3D phase-contrast flow measurements},
  author={Busch, Julia and Giese, Daniel and Wissmann, Lukas and Kozerke, Sebastian},
  journal={Magnetic resonance in medicine},
  volume={69},
  number={1},
  pages={200--210},
  year={2013},
  publisher={Wiley Online Library}
}

@article{ong_robust_2015,
  title={Robust 4D flow denoising using divergence-free wavelet transform},
  author={Ong, Frank and Uecker, Martin and Tariq, Umar and Hsiao, Albert and Alley, Marcus T and Vasanawala, Shreyas S and Lustig, Michael},
  journal={Magnetic resonance in medicine},
  volume={73},
  number={2},
  pages={828--842},
  year={2015},
  publisher={Wiley Online Library}
}

@article{ferdian_4dflownet_2020,
  title={4DFlowNet: super-resolution 4D flow MRI using deep learning and computational fluid dynamics},
  author={Ferdian, Edward and Suinesiaputra, Avan and Dubowitz, David J and Zhao, Debbie and Wang, Alan and Cowan, Brett and Young, Alistair A},
  journal={Frontiers in Physics},
  volume={8},
  pages={138},
  year={2020},
  publisher={Frontiers Media SA}
}

@article{fathi_super-resolution_2020,
  title={Super-resolution and denoising of 4D-flow MRI using physics-informed deep neural nets},
  author={Fathi, Mojtaba F and Perez-Raya, Isaac and Baghaie, Ahmadreza and Berg, Philipp and Janiga, Gabor and Arzani, Amirhossein and D’Souza, Roshan M},
  journal={Computer Methods and Programs in Biomedicine},
  volume={197},
  pages={105729},
  year={2020},
  publisher={Elsevier}
}

@article{donoho_adapting_1995,
  title={Adapting to unknown smoothness via wavelet shrinkage},
  author={Donoho, David L and Johnstone, Iain M},
  journal={Journal of the american statistical association},
  volume={90},
  number={432},
  pages={1200--1224},
  year={1995},
  publisher={Taylor \& Francis}
}

@article{garcia_role_2019,
  title={The role of imaging of flow patterns by 4D flow MRI in aortic stenosis},
  author={Garcia, Julio and Barker, Alex J and Markl, Michael},
  journal={JACC: Cardiovascular Imaging},
  volume={12},
  number={2},
  pages={252--266},
  year={2019},
  publisher={American College of Cardiology Foundation Washington, DC}
}

@article{hom_velocity-encoded_2008,
  title={Velocity-encoded cine MR imaging in aortic coarctation: functional assessment of hemodynamic events},
  author={Hom, Jeffrey J and Ordovas, Karen and Reddy, Gautham P},
  journal={Radiographics},
  volume={28},
  number={2},
  pages={407--416},
  year={2008},
  publisher={Radiological Society of North America}
}

@inproceedings{abdul-rahman_fast_2005,
  title={Fast three-dimensional phase-unwrapping algorithm based on sorting by reliability following a non-continuous path},
  author={Abdul-Rahman, Hussein and Gdeisat, Munther and Burton, David and Lalor, Michael},
  booktitle={Optical measurement systems for industrial inspection IV},
  volume={5856},
  pages={32--40},
  year={2005},
  organization={SPIE}
}

@article{bioucas-dias_phase_2007,
  title={Phase unwrapping via graph cuts},
  author={Bioucas-Dias, Jos M and Valadao, Gonalo},
  journal={IEEE Transactions on Image processing},
  volume={16},
  number={3},
  pages={698--709},
  year={2007},
  publisher={IEEE}
}

@article{loecher_phase_2016,
  title={Phase unwrapping in 4D MR flow with a 4D single-step laplacian algorithm},
  author={Loecher, Michael and Schrauben, Eric and Johnson, Kevin M and Wieben, Oliver},
  journal={Journal of Magnetic Resonance Imaging},
  volume={43},
  number={4},
  pages={833--842},
  year={2016},
  publisher={Wiley Online Library}
}

@article{yu2026review,
  title={Review of physics-informed neural networks in hemodynamics},
  author={Yu, Xianglong and Hu, Yu and Guo, Rui and Fan, Lei and Ding, Haiyan and Xiao, Jingjing},
  journal={Engineering Applications of Artificial Intelligence},
  volume={163},
  pages={112834},
  year={2026},
  publisher={Elsevier}
}

@article{richter2022neural,
  title={Neural conservation laws: A divergence-free perspective},
  author={Richter-Powell, Jack and Lipman, Yaron and Chen, Ricky TQ},
  journal={Advances in Neural Information Processing Systems},
  volume={35},
  pages={38075--38088},
  year={2022}
}

@article{garay2024physics,
  title={Physics-informed neural networks for parameter estimation in blood flow models},
  author={Garay, Jerem{\'\i}as and Dunstan, Jocelyn and Uribe, Sergio and Costabal, Francisco Sahli},
  journal={Computers in Biology and Medicine},
  volume={178},
  pages={108706},
  year={2024},
  publisher={Elsevier}
}

@inproceedings{hutter2014efficient,
  title={An efficient approach for assessing hyperparameter importance},
  author={Hutter, Frank and Hoos, Holger and Leyton-Brown, Kevin},
  booktitle={International conference on machine learning},
  pages={754--762},
  year={2014},
  organization={PMLR}
}

@article{rutkowski2021enhancement,
  title={Enhancement of cerebrovascular 4D flow MRI velocity fields using machine learning and computational fluid dynamics simulation data},
  author={Rutkowski, David R and Rold{\'a}n-Alzate, Alejandro and Johnson, Kevin M},
  journal={Scientific reports},
  volume={11},
  number={1},
  pages={10240},
  year={2021},
  publisher={Nature Publishing Group UK London}
}

@inproceedings{shone2023deep,
  title={Deep physics-informed super-resolution of cardiac 4D-flow MRI},
  author={Shone, Fergus and Ravikumar, Nishant and Lassila, Toni and MacRaild, Michael and Wang, Yongxing and Taylor, Zeike A and Jimack, Peter and Dall’Armellina, Erica and Frangi, Alejandro F},
  booktitle={International conference on information processing in medical imaging},
  pages={511--522},
  year={2023},
  organization={Springer}
}

@article{kalajahi2025input,
  title={Input parameterized physics informed neural networks for denoising, super-resolution, and imaging artifact mitigation in time resolved three dimensional phase-contrast magnetic resonance imaging},
  author={Kalajahi, Amin Pashaei and Csala, Hunor and Mamun, Zayeed Bin and Yadav, Sangeeta and Amili, Omid and Arzani, Amirhossein and D’Souza, Roshan M},
  journal={Engineering Applications of Artificial Intelligence},
  volume={150},
  pages={110600},
  year={2025},
  publisher={Elsevier}
}

@article{mella2026evaluating,
  title={Evaluating the impact of blood rheology in hemodynamic parameters by 4D flow MRI in large vessels considering the hematocrit effect},
  author={Mella, Hernan and Galarce, Felipe and Sekine, Tetsuro and Sotelo, Julio and Castillo, Ernesto},
  journal={Biomedical Signal Processing and Control},
  volume={111},
  pages={108145},
  year={2026},
  publisher={Elsevier}
}

@article{hansen1993use,
  title={The use of the L-curve in the regularization of discrete ill-posed problems},
  author={Hansen, Per Christian and O’Leary, Dianne Prost},
  journal={SIAM journal on scientific computing},
  volume={14},
  number={6},
  pages={1487--1503},
  year={1993},
  publisher={SIAM}
}

@article{singh2026vast,
  title={VAST: Vascular Flow Analysis and Segmentation for Intracranial 4D Flow MRI},
  author={Singh, Abhishek and Rayz, Vitaliy L and Vlachos, Pavlos P},
  journal={arXiv preprint arXiv:2601.13393},
  year={2026}
}

@article{hans2025smurf,
  title={SMURF: Scalable method for unsupervised reconstruction of flow in 4D flow MRI},
  author={Hans, Atharva and Singh, Abhishek and Vlachos, Pavlos and Bilionis, Ilias},
  journal={arXiv preprint arXiv:2505.12494},
  year={2025}
}

@article{carrillo_optimal_2019,
  title={Optimal dual-VENC unwrapping in phase-contrast MRI},
  author={Carrillo, Hugo and Osses, Axel and Uribe, Sergio and Bertoglio, Crist{\'o}bal},
  journal={IEEE transactions on medical imaging},
  volume={38},
  number={5},
  pages={1263--1270},
  year={2019},
  publisher={IEEE}
}

@article{hendrycks2016gaussian,
  title={Gaussian error linear units (gelus)},
  author={Hendrycks, Dan and Gimpel, Kevin},
  journal={arXiv preprint arXiv:1606.08415},
  year={2016}
}

@article{hertz2021sape,
  title={Sape: Spatially-adaptive progressive encoding for neural optimization},
  author={Hertz, Amir and Perel, Or and Giryes, Raja and Sorkine-Hornung, Olga and Cohen-Or, Daniel},
  journal={Advances in Neural Information Processing Systems},
  volume={34},
  pages={8820--8832},
  year={2021}
}

@article{kania2024fresh,
  title={Fresh: Frequency shifting for accelerated neural representation learning},
  author={Kania, Adam and Mihajlovic, Marko and Prokudin, Sergey and Tabor, Jacek and Spurek, Przemys{\'L} and others},
  journal={arXiv preprint arXiv:2410.05050},
  year={2024}
}

@article{iwata_measurement_2024,
	title = {Measurement of {Turbulent} {Kinetic} {Energy} in {Hypertrophic} {Cardiomyopathy} {Using} {Triple}-velocity {Encoding} {4D} {Flow} {MR} {Imaging}},
	volume = {23},
	number = {1},
	journal = {Magnetic Resonance in Medical Sciences},
	author = {Iwata, Kotomi and Sekine, Tetsuro and Matsuda, Junya and Tachi, Masaki and Imori, Yoichi and Amano, Yasuo and Ando, Takahiro and Obara, Makoto and Crelier, Gerard and Ogawa, Masashi and Takano, Hitoshi and Kumita, Shinichiro},
	year = {2024},
	pages = {39--48},
}

@inproceedings{tancik2021learned,
  title={Learned initializations for optimizing coordinate-based neural representations},
  author={Tancik, Matthew and Mildenhall, Ben and Wang, Terrance and Schmidt, Divi and Srinivasan, Pratul P and Barron, Jonathan T and Ng, Ren},
  booktitle={Proceedings of the IEEE/CVF conference on computer vision and pattern recognition},
  pages={2846--2855},
  year={2021}
}

@article{markl_comprehensive_2011,
  title={Comprehensive 4D velocity mapping of the heart and great vessels by cardiovascular magnetic resonance},
  author={Markl, Michael and Kilner, Philip J and Ebbers, Tino},
  journal={Journal of Cardiovascular Magnetic Resonance},
  volume={13},
  number={1},
  pages={7},
  year={2011},
  publisher={Elsevier}
}

@article{sotelo20163d,
  title={3D quantification of wall shear stress and oscillatory shear index using a finite-element method in 3D CINE PC-MRI data of the thoracic aorta},
  author={Sotelo, Julio and Urbina, Jesus and Valverde, Israel and Tejos, Cristian and Irarr{\'a}zaval, Pablo and Andia, Marcelo E and Uribe, Sergio and Hurtado, Daniel E},
  journal={IEEE transactions on medical imaging},
  volume={35},
  number={6},
  pages={1475--1487},
  year={2016},
  publisher={IEEE}
}

@article{tancik_fourier_2020,
  title={Fourier features let networks learn high frequency functions in low dimensional domains},
  author={Tancik, Matthew and Srinivasan, Pratul and Mildenhall, Ben and Fridovich-Keil, Sara and Raghavan, Nithin and Singhal, Utkarsh and Ramamoorthi, Ravi and Barron, Jonathan and Ng, Ren},
  journal={Advances in neural information processing systems},
  volume={33},
  pages={7537--7547},
  year={2020}
}

@article{jr_ladisa_computational_2011,
  title={Computational simulations demonstrate altered wall shear stress in aortic coarctation patients treated by resection with end-to-end anastomosis},
  author={Jr. LaDisa, John F and Dholakia, Ronak J and Figueroa, C Alberto and Vignon-Clementel, Irene E and Chan, Frandics P and Samyn, Margaret M and Cava, Joseph R and Taylor, Charles A and Feinstein, Jeffrey A},
  journal={Congenital heart disease},
  volume={6},
  number={5},
  pages={432--443},
  year={2011},
  publisher={Wiley Online Library}
}

@article{wilson_vascular_2013,
  title={The vascular model repository: a public resource of medical imaging data and blood flow simulation results},
  author={Wilson, Nathan M and Ortiz, Ana K and Johnson, Allison B},
  journal={Journal of medical devices},
  volume={7},
  number={4},
  pages={040923},
  year={2013},
  publisher={American Society of Mechanical Engineers}
}

@article{bissell_4d_2023,
  title={4D Flow cardiovascular magnetic resonance consensus statement: 2023 update},
  author={Bissell, Malenka and Raimondi, Francesca and Ait Ali, Lamia and Allen, Bradley D and Barker, Alex J and Bolger, Ann and Burris, Nicholas and Carh{\"a}ll, Carl-Johan and Collins, Jeremy D and Ebbers, Tino and others},
  journal={Journal of Cardiovascular Magnetic Resonance},
  volume={25},
  number={1},
  pages={40},
  year={2023},
  publisher={Elsevier}
}

@article{dirix_4dflow-unwrap_2024,
  title={4Dflow-unwrap: a collection of python-based phase unwrapping algorithms},
  author={Dirix, Pietro and Jacobs, Luuk and Kozerke, Sebastian},
  journal={International Society for Magnetic Resonance in Medicine, Singapore},
  year={2024}
}

@article{binter_bayesian_2013,
  title={Bayesian multipoint velocity encoding for concurrent flow and turbulence mapping},
  author={Binter, Christian and Knobloch, Verena and Manka, Robert and Sigfridsson, Andreas and Kozerke, Sebastian},
  journal={Magnetic resonance in medicine},
  volume={69},
  number={5},
  pages={1337--1345},
  year={2013},
  publisher={Wiley Online Library}
}

@article{takahashi2021four,
  title={Four-dimensional flow analysis reveals mechanism and impact of turbulent flow in the dissected aorta},
  author={Takahashi, Kenichiro and Sekine, Tetsuro and Miyagi, Yasuo and Shirai, Sayaka and Otsuka, Toshiaki and Kumita, Shinichiro and Ishii, Yosuke},
  journal={European Journal of Cardio-Thoracic Surgery},
  volume={60},
  number={5},
  pages={1064--1072},
  year={2021},
  publisher={Oxford University Press}
}

@article{saitta_implicit_2024,
  title={Implicit neural representations for unsupervised super-resolution and denoising of 4D flow MRI},
  author={Saitta, Simone and Carioni, Marcello and Mukherjee, Subhadip and Sch{\"o}nlieb, Carola-Bibiane and Redaelli, Alberto},
  journal={Computer methods and programs in biomedicine},
  volume={246},
  pages={108057},
  year={2024},
  publisher={Elsevier}
}



\end{document}